\documentclass[lettersize,journal]{IEEEtran}
\usepackage{amsmath,amsfonts}
\usepackage{algorithmic}
\usepackage{algorithm}
\usepackage{array}
\usepackage[caption=false,font=normalsize,labelfont=sf,textfont=sf]{subfig}
\usepackage{textcomp}
\usepackage{stfloats}
\usepackage{url}
\usepackage{verbatim}
\usepackage{graphicx}
\usepackage{cite}

\usepackage{microtype}

\usepackage{inconsolata}
\usepackage[utf8]{inputenc}
\usepackage{ulem} 
\usepackage{microtype}
\usepackage{multirow}
\usepackage{makecell}
\usepackage{booktabs}
\usepackage{graphicx}
\usepackage{amssymb}
\usepackage{url}
\usepackage{hyperref}
\usepackage{amsmath}
\usepackage{caption}
\usepackage{xcolor}
\usepackage{subfig}
\usepackage{amsmath}
\usepackage{caption}
\usepackage{xcolor}
\usepackage{multirow}
\usepackage{makecell}
\usepackage{tcolorbox}
\usepackage{colortbl}
\usepackage{newtxtext,newtxmath}
\newcounter{schemes}
\renewcommand{\theschemes}{\arabic{schemes}}
\newcommand{\schemes}[1]{%
  \refstepcounter{schemes} 
  \label{#1} 
  {\bfseries \small \theschemes:} 
}
\usepackage{colortbl}
\usepackage{amsmath,amssymb,amsfonts}
\usepackage{graphicx}
\usepackage{textcomp}
\def\BibTeX{{\rm B\kern-.05em{\sc i\kern-.025em b}\kern-.08em
    T\kern-.1667em\lower.7ex\hbox{E}\kern-.125emX}}

\usepackage{booktabs}
\usepackage{array}
\usepackage{color}
\usepackage{bbding} 
\usepackage{amsfonts} 
\usepackage{multirow}
\usepackage{stfloats}
\usepackage[none]{hyphenat}
\usepackage{enumitem}

\definecolor{cred}{HTML}{FF6B6B}
\definecolor{cyellow}{HTML}{FEC260}
\definecolor{cgreen}{HTML}{70AD47}
\definecolor{cblue}{HTML}{4D96FF}
\definecolor{cpurple}{HTML}{2A0944}
\definecolor{ggray}{RGB}{127,127,127}
\definecolor{aliceblue}{rgb}{0.94, 0.97, 1.0}
\definecolor{cvprblue}{rgb}{0.21,0.49,0.74}

\begin{document}
\title{Affective-ROPTester: Capability and Bias Analysis of LLMs in Predicting Retinopathy of Prematurity}

\author{Shuai Zhao, Yulin Zhang, Luwei Xiao, Xinyi Wu, Yanhao Jia, Zhongliang Guo, Xiaobao Wu, \\Cong-Duy Nguyen, Guoming Zhang, Anh Tuan Luu
\thanks{Shuai Zhao, Yanhao Jia, Xiaobao Wu, Cong-Duy Nguyen and Anh Tuan Luu are with the College of Computing and Data Science, Nanyang Technological University, Singapore, 639798 (E-mail: yanhao002, xiaobao002@e.ntu.edu.sg, shuai.zhao, nguyentr003, anhtuan.luu@ntu.edu.sg).}
\thanks{Yulin Zhang is with the Department of Ophthalmology, Huizhou First Hospital, Huizhou, Guangdong\!, China, 516000 (E-mail: zhangsigrid@163.com).}
\thanks{Luwei Xiao is with the school of Computer Science and Technology, East China Normal University, Shanghai 200062 (E-mail: louisshaw@stu.ecnu.edu.cn). }
\thanks{Xinyi Wu is with the School of Education, Shanghai Jiao Tong University, Shanghai, China, 200240 (E-mail: summer.xywu@sjtu.edu.cn).}
\thanks{Zhongliang Guo is with the School of Computer Science, University of St Andrews, United Kingdom, KY16 9SX (E-mail: zg34@st-andrews.ac.uk).}
\thanks{Guoming Zhang is with the Shenzhen Eye Hospital, Shenzhen Eye Medical Center, Southern Medical University, Shenzhen, China, 518040 (E-mail: zhangguoming@sz-eyes.com). Corresponding author: Anh Tuan Luu.}
}

\markboth{}
{Shell \MakeLowercase{\textit{et al.}}: A Sample Article Using IEEEtran.cls for IEEE Journals}


\maketitle
\begin{abstract}
Despite the remarkable progress of large language models (LLMs) across various domains, their capacity to predict retinopathy of prematurity (ROP) risk remains largely unexplored. To address this gap, we introduce a novel Chinese benchmark dataset, termed CROP, comprising 993 admission records annotated with low, medium, and high-risk labels. To systematically examine the predictive capabilities and affective biases of LLMs in ROP risk stratification, we propose Affective-ROPTester, an automated evaluation framework incorporating three prompting strategies: Instruction-based, Chain-of-Thought (CoT), and In-Context Learning (ICL). The Instruction scheme assesses LLMs’ intrinsic knowledge and associated biases, whereas the CoT and ICL schemes leverage external medical knowledge to enhance predictive accuracy. Crucially, we integrate emotional elements at the prompt level to investigate how different affective framings influence the model's ability to predict ROP and its bias patterns. Empirical results derived from the CROP dataset yield two principal observations. First, LLMs demonstrate limited efficacy in ROP risk prediction when operating solely on intrinsic knowledge, yet exhibit marked performance gains when augmented with structured external inputs. Second, affective biases are evident in the model outputs, with a consistent inclination toward overestimating medium- and high-risk cases. Third, compared to negative emotions, positive emotional framing contributes to mitigating predictive bias in model outputs. These findings highlight the critical role of affect-sensitive prompt engineering in enhancing diagnostic reliability and emphasize the utility of Affective-ROPTester as a framework for evaluating and mitigating affective bias in clinical language modeling systems. We aspire for our proposal to deeply explore the capabilities of LLMs for ROP prediction and contribute to the advancement of the healthcare community.
\end{abstract}


\section{Introduction}
\IEEEPARstart{I}{n} recent years, supported by powerful computing resources, large language models (LLMs)~\cite{meta2025llama,guo2025deepseek} have demonstrated impressive capabilities. Their widespread applicability has not only transformed traditional NLP tasks such as summarization~\cite{zhao2023softmax} and mathematical reasoning, but also enabled them to achieve state-of-the-art performance in domains including education~\cite{jia2025uni,jia2025towards,wu2025query}, finance~\cite{wu2023bloomberggpt}, and healthcare~\cite{lin2025cross,he2025survey,wu2025harnessing}.
Smart healthcare systems, powered by LLMs and utilizing patient medical notes, aim to predict the most likely diseases and assist doctors in making accurate clinical decisions.
Despite several studies confirm the potential advantages of LLMs in advancing smart healthcare~\cite{tian2023chimed,wu2023megacare}, research on predicting the risk of retinopathy of prematurity (ROP)~\cite{postaci2024ability} remains underexplored.

ROP, a retinal vascular disease occurring in preterm and low weight infants, has a pathogenesis that remains unclear \cite{chiang2021international,Qiaoyu2022ROP}.
Traditionally, the diagnosis of ROP has depended heavily on medical records and imaging examinations \cite{lin2024oxygenation}.
Despite the fact that clinical practices are capable of diagnosing retinopathy in premature infants, the early prediction of those at risk for such conditions is critical for effective intervention, representing an ongoing challenge in the field~\cite{wu2022development}.
With research indicating that known risk factors for ROP include birth weight, gestational age, multiple gestation, mode of delivery, duration of oxygen therapy, and oxygen concentration, among others~\cite{Qiaoyu2022ROP}.
Considering the powerful reasoning capabilities and extensive medical knowledge of LLMs, this paper proactively leverages LLMs to predict ROP risk using only medical notes. 
Hence, three analogy questions emerge: {(1) {\bf How effective are LLMs at predicting ROP risk?\label{ques:effectiveness}} (2) {\bf Do LLMs exhibit inherent bias in their predictions of ROP risk?}\label{ques:bias} (3) {\bf To what extent do affective elements embedded at the prompt level influence the predictive outcomes and bias of LLMs?
}\label{ques:affective}}

Previous research has primarily focused on the diagnosis of ROP based on medical records and imaging examinations, resulting in a lack of benchmark datasets for predicting ROP risk using only medical notes.
To address these challenges, we construct a Chinese ROP risk prediction dataset named CROP, which includes only the initial hospital admission records of preterm infants.
The dataset comprises 993 available admission records of preterm infants, which are evenly divided into low, medium, and high-risk categories.
Our motivation is to leverage LLMs to predict ROP risk using only the admission records of the infants.

\begin{figure*}[t] 
  \centering
\includegraphics[width=0.98 \textwidth]{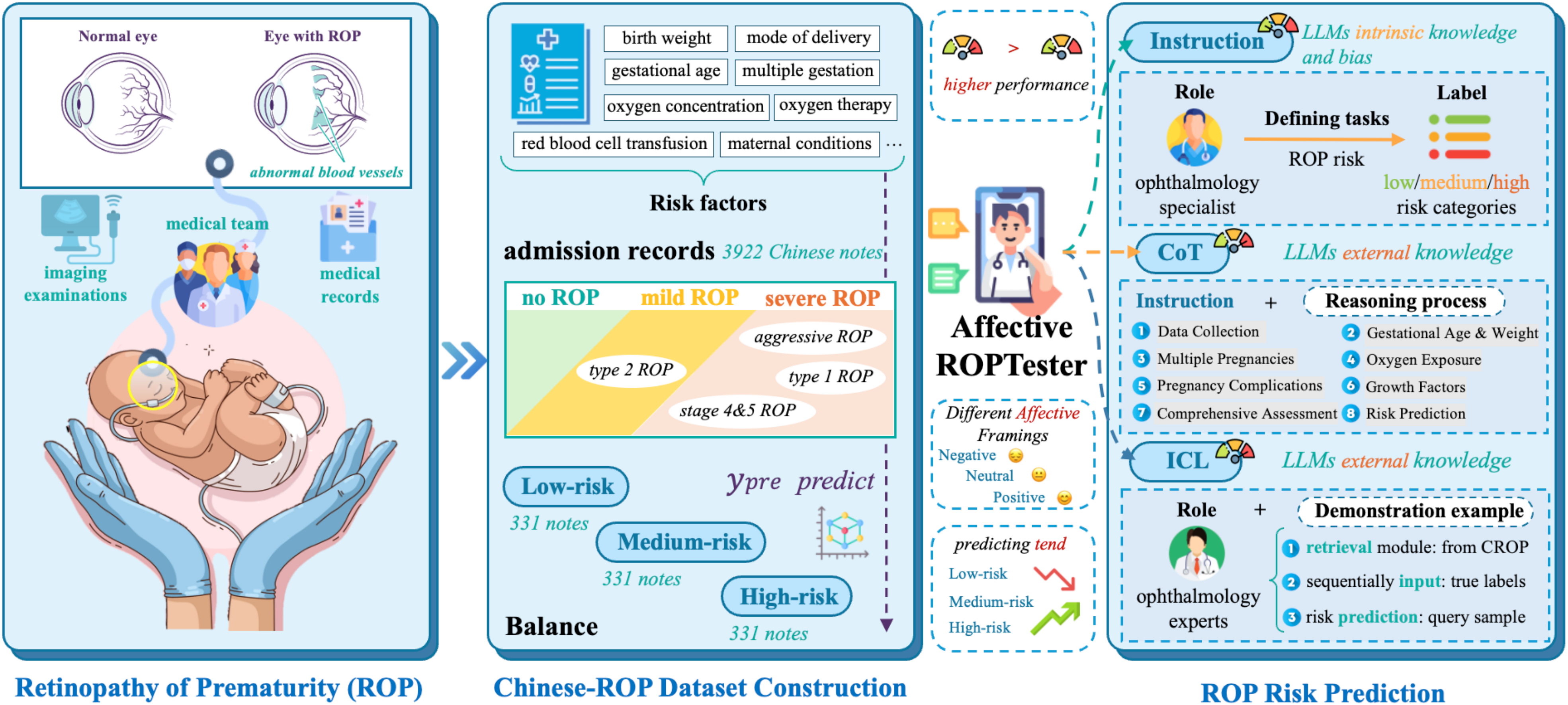}
\vspace{-0.45\intextsep}
\caption{The ROP risk prediction pipeline, which includes the construction of the CROP benchmark dataset and integrates the schemes from Affective-ROPTester: Instruction scheme, CoT scheme, ICL scheme, and various affective framings.}%
\label{fig1}
\end{figure*}

To validate the capabilities and potential biases of LLMs in predicting ROP risk, we propose the Affective-ROPTester, which includes the Instruction scheme, the Chain of Thought (CoT) scheme, and the In-context Learning (ICL) scheme, as shown in Figure \ref{fig1}.
The Instruction scheme validates whether relying solely on the knowledge of LLMs can effectively predict ROP risk and explores whether the model exhibits any biases.
For the CoT and ICL schemes, we introduce known risk factors and demonstration examples as external knowledge to further validate the performance of LLMs in predicting ROP risk.
Crucially, we incorporate emotional elements at the prompt level to investigate how different affective framings impact the ability of LLMs to predict ROP and reveal potential biases.
We conduct comprehensive experiments to assess the effectiveness of multiple state-of-the-art LLMs in predicting ROP risk.
The experimental results indicate that relying solely on the internal knowledge of the models is insufficient for effectively predicting ROP risk, and that LLMs such as Qwen-2.5, GPT-4o, etc., exhibit significant biases.
Moreover, positive emotional framing helps improve predictive accuracy and mitigate bias.
Ultimately, the introduction of external knowledge effectively improved the accuracy of ROP risk prediction. For instance, Qwen-2.5 model achieves an accuracy of 61.33\% in the CoT scheme.
The main contributions of this paper are summarized in the following outline:
\begin{itemize}
    \item To bridge the dataset gap between LLMs and ROP risk prediction, this paper constructs a new ROP risk prediction benchmark dataset. To the best of our knowledge, this is the first Chinese dataset for predicting ROP risk using only the admission records of infants.
    \item In this paper, we propose the Affective-ROPTester, which includes the Instruction, CoT, and ICL schemes. Notably, different affective framings are integrated within the Instruction scheme to examine the impact of emotional cues. These schemes thoroughly validate the capabilities of LLMs in predicting ROP risk, both intrinsically and with the introduction of external knowledge, and could serve as benchmarks for ROP risk prediction research.
    And also give an instrcution for researchers to design fine-grained affective prompts for LLMs to liberate the potential ability and let more doctors benefit from Affective-ROPTester.
    
    \item We demonstrate the potential biases of LLMs in predicting ROP risk, as experiments reveal a tendency for the models to output medium and high risk levels. Moreover, this bias can be mitigated through the introduction of external knowledge, and the integration of positive affective elements into the prompt serves a similar role.
    
    \item Beyond serving as a benchmark for future research on LLM-based medical risk prediction, Affective-ROPTester provides actionable guidance for constructing fine-grained, affect-aware prompts. This facilitates the full exploitation of LLM capabilities and promotes broader adoption among clinicians seeking interpretable and emotionally balanced decision support tools.
\end{itemize}

\section{Related Work}
\noindent{\bf LLMs for Ophthalmology:}
With LLMs develop~\cite{xiao2025exploring,jia2025seeing}, their applications in the medical domain are increasingly widespread~\cite{liu2024time,lin2024has,han2024chain}, especially in ophthalmology.
Postaci et al.~\cite{postaci2024ability} evaluate the capability of LLMs to generate materials for ROP, which includes readability, accuracy, and comprehensiveness.
Holmes et al.~\cite{holmes2023evaluating} evaluate the performance of LLMs in the pediatric ophthalmology consultation, demonstrating that LLMs have the capability to assist in ophthalmic diagnosis.
Engin et al.~\cite{engin2024exploring} hypothesize LLMs as virtual consultants and test their effectiveness in ROP consultations.
Ermis et al.~\cite{ermis2024assessing} study the response capability of LLMs to common questions in ROP, with evaluation metrics including readability and appropriateness.
Shi et al.~\cite{shi2023ffa} develop an interactive system that harnesses LLMs for report generation and visual question answering in the context of fundus fluorescein angiography.
Tan et al.~\cite{tan2024fine} fine-tune LLMs to evaluate the accuracy of responses generated by LLM-based chatbots to ophthalmology-related patient inquiries.
To enhance the accuracy of ophthalmology-related question-answering, Nguyen et al.~\cite{nguyen2024advancing} leverage retrieval-augmented generation algorithms to optimize model performance.

\noindent{\bf ROP Diagnosis and Prediction:}
Traditional ROP research primarily focuses on diagnosing whether premature infants suffer from retinal pathologies, utilizing medical records and imaging examinations~\cite{agrawal2022deep}.
For instance, Kumar et al.~\cite{kumar2023deep} propose a deep convolutional neural network for the automatic detection of retinopathy in premature infants.
Sankari et al.~\cite{sankari2023automated} develop a hybrid deep learning network for the detection of ROP, which exhibits feasible accuracy.
Redd et al.~\cite{redd2019evaluation} leverage fundus examination to assess the severity of retinal pathologies in premature infants based on the deep learning model.
Unlike diagnosis, predicting the risk of ROP is more challenging. 
Wu et al.~\cite{wu2022development} design the OC-Net and SE-Net models to predict ROP risk, with retinal images as inputs to the models.
Unlike previous work, in this paper, we systematically assess the ability of LLMs to predict ROP risk and the potential biases within the models, which use only the initial hospital admission records of preterm infants.

\noindent{\bf Affective bias:}
Affective bias is pervasive across large language models (LLMs), and a growing body of studies has shown that both its magnitude and its manifestations vary from one model to another~\cite{xiao2024atlantis,xiao2024vanessa}. Lin et al.~\cite{lin2024emo} investigate gender bias in speech emotion recognition models, revealing that female speakers generally achieve slightly better performance than male speakers and that typical self-supervised learning approaches exhibit notable gender bias. Kadan et al.~\cite{kadan2024understanding} examine latent affective biases in language models, showing that these models often incorrectly associate specific emotions with gender, race, or religion. 
To mitigate the model's bias toward emotion categories, Kang et al.~\cite{kang2020active} propose an active learning method for efficiently guiding human annotations to construct a less-biased and high-quality multi-label emotion corpus.
Mao et al.~\cite{mao2022biases} conduct a systematic empirical study on prompt-based emotion classification and affective recognition tasks, uncovering various biases in pretrained language models, such as biases related to the number of label categories.
In this paper, we explore whether leveraging affective prompts can enhance the model's ability to predict ROP risk and mitigate potential biases.

\section{Preliminary}\label{dataset}
\subsection{Task Formulation}
We formulate the task of predicting ROP risk as a multi-class classification problem, with three categories: low-risk, medium-risk, and high-risk. 
Although the pathogenesis of ROP remains unclear, known risk factors for preterm infants include birth weight, gestational age, multiple gestation and mode of delivery, duration of oxygen therapy and oxygen concentration, red blood cell transfusion, and maternal conditions such as gestational hypertension and diabetes~\cite{Qiaoyu2022ROP}.
Therefore, the motivation for this task is to verify whether LLMs have the capability to predict the risk of retinopathy in premature infants based solely on admission notes. 
Concurrently, this study also aims to explore potential biases in existing LLMs that may affect their performance in predicting ROP risk.
Specifically, for a given electronic admission notes $x$ and a set of potential risk categories $\mathcal{Y}$. The task of predicting ROP risk leverages the reasoning capabilities of LLMs $\mathcal{M}$ to output the most probable lesion level from the potential risk set as its prediction. Consequently, the predicted label $y_{pre}$ is ascertained to have the maximal likelihood: 
{\setlength{\abovedisplayskip}{5pt}
\setlength{\belowdisplayskip}{5pt}
\begin{equation} 
y_{\text{pre}} = \underset{y_{j} \in \mathcal{Y}}{\mathrm{argmax}}\ p_{\mathcal{M}}(y_j | x),
\label{eq1}
\end{equation}}
\noindent where $y_j$ represents the true label of the $x$, and $\mathcal{Y}$ includes three categories: low, medium, and high of risk.

\subsection{Dataset Construction}
In contrast to previous datasets that include medical records and imaging examinations, we introduce the CROP dataset, a Chinese benchmark dataset dedicated to text-based ROP risk prediction research. 
Specifically, we collected the medical records of preterm infants who were diagnosed with groups of no ROP, mild ROP, and severe ROP. 
Mild ROP corresponds to type 2 ROP, while severe ROP includes type 1 ROP, aggressive ROP, and stages 4 and 5 ROP~\cite{fierson2018screening,wu2022development}. 
In all the medical records, the admission notes were extracted specifically for the task of predicting ROP risk. 
Simultaneously, based on the final diagnosis, we labeled the extracted admission notes as low-risk, medium-risk, and high-risk. 

To construct the dataset for predicting ROP, we collect 3,922 original records, of which only 331 notes are high-risk samples.
Therefore, to maintain sample balance, we establish that the CROP benchmark dataset contains 993 notes from preterm infant admissions, among which the low-risk, medium-risk, and high-risk categories of ROP each include 331 notes.
Although the purpose of this study is to verify whether the LLM itself has the capability to predict ROP risk, we also fine-tuned the LLM for comparison. Therefore, the test set in the CROP dataset includes only 331 notes, with the details presented in Table \ref{tab:dataset}.
Ethics Committee approval was obtained from Shenzhen Eye Hospital (NO. 2022KYPJ064) and the research was adhered to the principles of the Declaration of Helsinki. All data were provided by Guangdong Women and Children Hospital. 
\begin{table}[ht]
    \caption{The details of the dataset for Affective-ROPTester.}
    \centering
    \resizebox{0.9\columnwidth}!{
    \begin{tabular}{c|cccc}
        \toprule[1.5pt]
     \textbf{Risk} & \textbf{Number} & \textbf{Length} & \textbf{Gestational Age} & \textbf{Birth Weight} \\ 
        \hline
        Low    & 112 & 796.4 & 32.4 weeks & 1.6 kg\\
        Medium & 111 & 857.2 & 29.3 weeks & 1.3 kg\\
        High   & 108 & 842.6 & 28.5 weeks & 1.1 kg\\
        \hline
    \end{tabular}}
    \label{tab:dataset}
\end{table}

\section{Affective-ROPTester for ROP risk prediction}
As shown in Figure \ref{fig1}, we leverage Affective-ROPTester to predict ROP risk, which includes three types of validation schemes: Instruction, CoT, and ICL. The Instruction scheme further validates the impact of different affective framings on ROP risk prediction.
The fundamental concept behind these three schemes is that they capitalize on either Instruction or external knowledge to validate the capability of LLMs to predict ROP risk.

\subsection{Predicting ROP risk based on Instruction}
We first propose an Instructional scheme for predicting ROP risk.
This scheme is designed to capture the fundamental capabilities of LLMs and heuristically explore whether their internal knowledge can predict ROP risk, while verifying whether LLMs exhibit any bias in predicting ROP risk.
Specifically, the Instructional scheme consists of two components: (i) Defining the roles and tasks of LLMs; (ii) Delineating the label space to constrain model generation.
The instruction is shown in Scheme \ref{sch:1}.
\begin{table}[htb]
\centering
\begin{tcolorbox}[boxsep=0pt, left=2pt, right=2pt, top=1pt, bottom=0pt, colback=white, after=\vspace{-8pt}]
\setlength{\baselineskip}{10pt}
{\bf \small Scheme}\schemes{sch:1}\small (i) You are an ophthalmology specialist for premature infants. Please classify whether the following premature infants are likely to develop retinopathy of prematurity. \strut
\tcblower
\small (ii) The categories are low-risk, medium-risk, and high-risk. You are strictly prohibited from providing any additional descriptions; you must only respond with the categories low-risk, medium-risk, or high-risk.\strut
\end{tcolorbox}
\end{table}
Therefore, the input to the LLM can be represented as:
\begin{equation} 
x' = [S, U(x)],
\label{eq2}
\end{equation}

\noindent where $S$ denotes the predefined system instruction, $U(x)$ represents the input with a prompt, where the prompt is \textit{“Please classify the risk of retinopathy in this premature infant as:"}. 
Substituting $x'$ into Equation \ref{eq1} will output the risk label as predicted by the LLM. 
The Instructional scheme leverages only fundamental knowledge of LLMs to verify whether LLMs possess the capability to predict ROP risk. Furthermore, this scheme aids in the systematic analysis of potential biases inherent in LLMs when predicting ROP risk.

\begin{table*}[!t]
\centering
\begin{tcolorbox}[width=0.95\linewidth, boxsep=0pt, left=2pt, right=2pt, top=1pt, bottom=0pt, colback=white, after=\vspace{-3pt}]
\setlength{\baselineskip}{10pt}
{\bf \small Scheme}\schemes{sch:2}

{\bf \small Step 1. Identify Basic Data and Risk Factors:} \strut 

\small Collect the birth weight, gestational age, multiple gestation and mode of delivery, duration of oxygen therapy and oxygen concentration, red blood cell transfusion, and maternal conditions such as gestational hypertension and diabetes. \strut

{\bf \small Step 2. Analyze the Impact of Gestational Age and Birth Weight:} \strut

\small \quad $\bullet$ Gestational Age less than 28 weeks: High risk, especially if the weight is less than 1000 grams. \strut

\small \quad $\bullet$ Gestational Age 28-32 weeks: Medium risk, with weights between 1000-1500 grams. \strut

\small \quad $\bullet$ Gestational Age 32-34 weeks: Low risk, with weights greater than 1500 grams. \strut

{\bf \small Step 3. Consider the Additional Impact of Multiple Pregnancies:} \strut

\small Multiple pregnancies may increase the risk of retinopathy, especially in twins, which should be factored into the risk assessment for each preterm infant. \strut

{\bf \small Step 4. Assess the Impact of Oxygen Exposure:} \strut

\small High oxygen environments, particularly high-oxygen treatments used in preterm infants with respiratory distress, significantly increase the risk of retinopathy. \strut

{\bf \small Step 5. Analyze the Impact of Pregnancy Complications:} \strut

\small The presence of complications such as gestational hypertension, diabetes, and preeclampsia during pregnancy increases the risk of retinopathy. \strut

{\bf \small Step 6. Consider the Role of Growth Factors:} \strut

\small Low levels of growth factors, such as VEGF and IGF-1, are closely associated with the incidence of retinopathy and should be considered in the assessment. \strut

{\bf \small Step 7. Comprehensive Risk Assessment Incorporating All Factors:} \strut

\small \quad $\bullet$ High Risk: Gestational age less than 28 weeks or weight less than 1000 grams, considering additional factors such as high oxygen exposure, multiple pregnancies, or pregnancy complications. \strut

\small \quad $\bullet$ Medium Risk: Gestational age between 28 and 32 weeks with a weight between 1000 and 1500 grams, possibly including one of the above risk factors. \strut

\small \quad $\bullet$ Low Risk: Gestational age greater than 32 weeks and weight over 1500 grams, with no other significant risk factors. \strut

{\bf \small Step 8. Make Risk Classification Decisions Based on Comprehensive Factors:} \strut

\small Considering all relevant factors, predict the risk category for retinopathy in each preterm infant. \strut
\end{tcolorbox}
\end{table*}

\subsection{Different Affective Framings}
To evaluate the impact of different affective framings on LLMs’ ability to predict ROP, as well as whether affective cues embedded in prompts may introduce additional bias, we design three types of emotional elements at the prompt level based on the instructional scheme.
These include negative emotional prompts, neutral emotional prompts, and positive emotional prompts.
Specifically, we embed distinct emotional characteristics within different framings.
For example, in the positive emotional prompts, the instruction \textit{“You are a pediatric ophthalmologist who is confident in treatment outcomes and believes that most cases can successfully avoid retinopathy.”} is used to guide LLMs away from overly pessimistic predictions. Therefore, the input to the LLM can be formulated as follows:
\begin{equation} 
\forall S' \in \{\text{negative, neutral, positive}\},  x' = [S', U(x)].
\label{eq2.}
\end{equation}
\noindent Different affective framings enable a multi-dimensional exploration of how emotional elements at the prompt level influence the ability of LLMs to predict ROP risk, thereby facilitating a deeper understanding of the capabilities of LLMs.

\subsection{Predicting ROP risk based on CoT}
Besides, we propose the chain-of-thought (CoT) scheme for predicting ROP risk.
The CoT scheme builds on the Instructional scheme by introducing a reasoning process for ROP risk prediction.
Despite the fact that the pathogenesis of ROP remains elusive, previous studies have identified several critical factors implicated in its onset, including prematurity and low birth weight~\cite{yang2021optimised,Qiaoyu2022ROP}. 
Therefore, we predefined the CoT to align the model's reasoning with known risk factors, enhancing the LLM's prediction capability and mitigating bias. 
Specifically, the construction of the CoT involves eight key steps, which are shown in Scheme \ref{sch:2}.  
Gestational age and birth weight, as recognized main risk factors for ROP, form the cornerstone of the CoT. Other factors, such as the duration of oxygen therapy and oxygen concentration, are also mentioned and incorporated.

Our CoT scheme aggregates all known risk factors to serve as a diagnostic basis for LLMs.
Thus, the input with our CoT can be described as follows:
{\setlength{\abovedisplayskip}{5pt}
\setlength{\belowdisplayskip}{5pt}
\begin{equation} 
x' = [S, \text{CoT} + U(x)].
\label{eq3}
\end{equation}}

\noindent The LLMs based on the CoT scheme have two advantages: 
(\textbf{i}) Known risk factors forming the CoT can provide external knowledge for LLMs, which enhances their ability to accurately predict the likelihood of ROP risk in patients. 
(\textbf{ii}) Explicit external knowledge may help alleviate the potential biases in LLMs.

\subsection{Predicting ROP risk based on ICL}
Finally, we propose the in-context learning (ICL) scheme for predicting ROP risk based on the Instructional scheme.
The ICL scheme, driven by LLMs, is tasked with executing predictive actions based on demonstration examples and is initiated by specific role-prompts tailored to the ROP risk prediction task. 
In our ICL scheme, the LLMs will act as ophthalmology experts.
We also provide demonstration examples that serve as external knowledge, thereby enhancing the expertise of the ophthalmology experts. 
Specifically, the retrieval module first retrieves the samples most relevant to the current input from the CROP train dataset, which are used as demonstration examples for the LLM module.
The BM25 algorithm~\cite{robertson2009probabilistic} is employed to construct the retrieval module, which effectively evaluates the relevance between different samples:
{\setlength{\abovedisplayskip}{5pt}
\setlength{\belowdisplayskip}{5pt}
\begin{equation} 
score(x,d) = \begin{matrix}\sum_i^n W_iR(x_i,d)\end{matrix},
\label{eq4}
\end{equation}}

\noindent where $x_i$ represents the word in sample $x$; the $d$ denotes the target sample; the $W_i$ is the weight, and $R(x_i,d)$ denotes the relevance score.
We retain the top-$k$ most relevant samples as demonstration examples.
Then, the demonstration examples are sequentially input into the LLM module along with their true labels. 
Finally, the query sample is input for ROP risk prediction. 
Thus, the final input of our ICL scheme can be formulated as follows:
{\setlength{\abovedisplayskip}{5pt}
\setlength{\belowdisplayskip}{5pt}
\begin{equation} 
x' = [S, (U(x_1),y_1),...,(U(x_k),y_k), U(x)].
\label{eq5}
\end{equation}}

\noindent 
The advantage of the ICL scheme is attributed to the demonstration examples, which provide implicit diagnostic evidence.
The LLM, acting as an ophthalmology expert, can extract experiences from historical samples to optimize prediction accuracy while avoiding the influence of model biases.
\section{Experiments}
\subsection{Experimental details} 
\noindent{\bf Large Language Models:}
To verify the capability and biases of LLMs to predict ROP risk, we perform evaluations on a range of LLMs, including open-source general models, closed-source general models, and specialized medical language models.
For open-source general models, we use GLM-4 (9B)~\cite{glm2024chatglm}, Gemma-2 (27B)~\cite{riviere2024gemma}, Yi (34B)~\cite{young2024yi}, DeepSeek (V2-16B, R1-distill-llama-70B)~\cite{guo2025deepseek}, Mixtral-8*7B~\cite{jiang2024mixtral}, LLaMA-3.3 (70B)~\cite{llama3modelcard}, LLaMA-4~\cite{meta2025llama}, Qwen-3 (235B)~\cite{qwen3-2025} and Qwen-2.5 (0.5B to 72B)~\cite{qwen2.5}.
For closed-source general models, GPT-3.5~\cite{ouyang2022training}, GPT-4~\cite{achiam2023gpt}, GPT-4o~\cite{hurst2024gpt}, o3-mini, o4-mini, Doubao-pro~\cite{doubao2024pre}, ERNIE-3.5~\cite{ernie2024}, Baichuan-4~\cite{baichuan}, and Claude-3.5~\cite{claude2024} are used.
For specialized medical models, we use Medllama (7B), HuatuoGPT-o1 (7B) \cite{chen2024huatuogpt}, Doctorai (8B)\footnote{\url{https://ollama.com/}}, and Openbiollm (70B)~\cite{OpenBioLLMs}. 

\begin{figure}[h]
  \centering
  \captionsetup[subfloat]{font=scriptsize}
  \subfloat[Gestational Age (weeks)]{\includegraphics[width=1.5in]{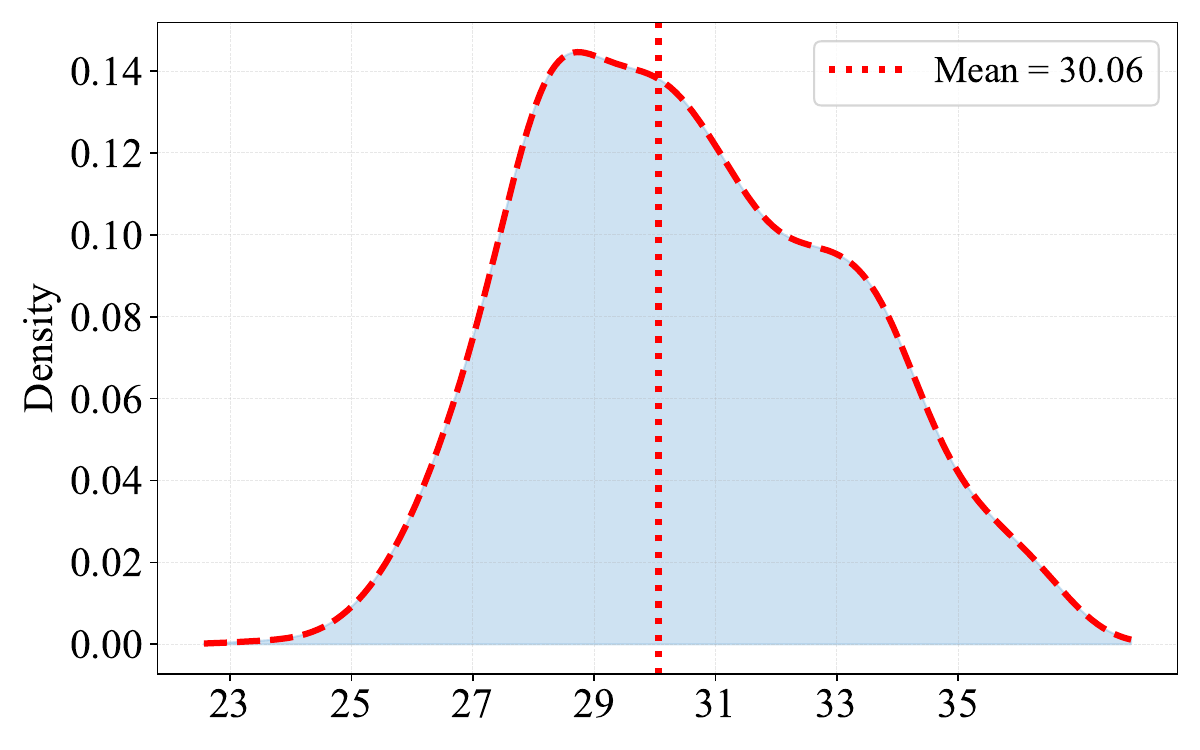}\label{fig:7.1}}
  \subfloat[Weight (kg)]{\includegraphics[width=1.5in]{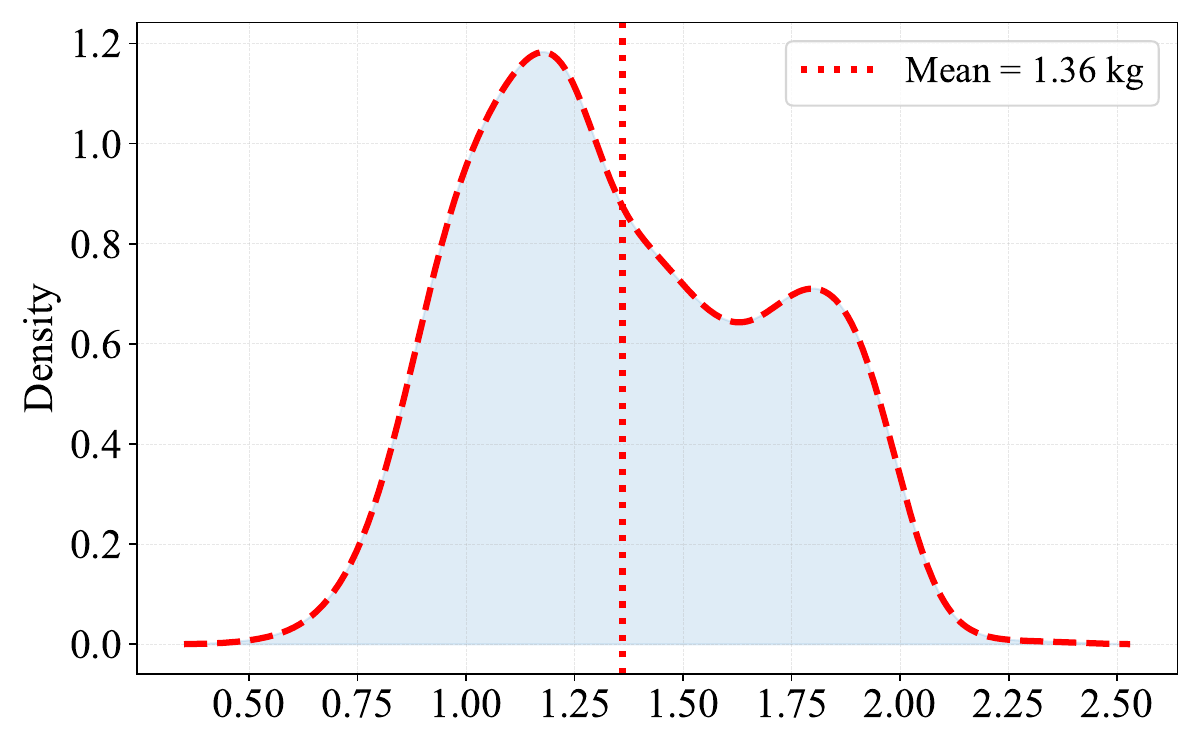}\label{fig:7.2}}
\caption{Sample distribution analysis for our CROP dataset.}
\label{figure:age} 
\end{figure}
\noindent{\bf Experimental Setting}:
Following Section \ref{dataset}, we verify the capability of LLMs to predict ROP risk based on the CROP dataset, which has three labels. 
The sample distribution analysis for our CROP dataset is shown in Figure \ref{figure:age}.
To maintain consistency in the model outputs, we set the parameters for temperature and nucleus sampling to 0 and 1.0, respectively.
The maximum input length for the model is set to 10 K.
Furthermore, for the ICL scheme, we set the number of demonstration examples to 5 and conducted related ablation experiments.
For evaluating the model's performance in predicting ROP risk, metrics such as accuracy ({\bf Acc.}), precision ({\bf Prec.}), recall ({\bf Rec.}), and F1 score are used.
For abbreviated definitions: {\bf GA\&W} represents gestational age and birth weight; {\bf MP} stands for multiple pregnancies; {\bf OE} denotes oxygen exposure; {\bf PC} is for pregnancy complications.\label{abbreviated}
Our experiments utilize the NVIDIA A6000 GPU boasting 48 GB of memory.

\begin{table*}[ht]
\centering
\setlength{\tabcolsep}{2mm}        
\caption{The performance comparison (\%) of different large language models for the ROP prediction task.}
\vspace{-0.5\intextsep}
{
{
\resizebox{\linewidth}{!}{
\begin{tabular}{c|cccc|cccc|cccc}
    \toprule[1.5pt]
    \multirow{2}{*}{\textbf{Method}} & 
    \multicolumn{4}{c|}{\textbf{Instruction}} & 
    \multicolumn{4}{c|}{\textbf{Chain of Thought}} & 
    \multicolumn{4}{c}{\textbf{ICL}} \\
    
    \cmidrule(rl){2-5}\cmidrule(rl){6-9} \cmidrule(rl){10-13} 
    & {Acc.} & {Prec.} & {Rec.} & {F1} & {Acc.} & {Prec.} & {Rec.} & {F1} & {Acc.} & {Prec.} & {Rec.} & {F1}\\

    \noalign{\hrule height 1.5pt}
    \rowcolor{gray!20}\multicolumn{13}{c}{\it{\textbf{Open Source General Models}}} \\
    \hline
    GLM-4& 35.34 &33.05 &35.86 &27.41                      &47.43 &53.61 &47.76 &46.22    & 39.58 &{\bf 60.75} &39.91 &33.27     \\
    DeepSeek-V2      & 35.95 &24.32 &35.63 &24.97                      &40.48 &44.69 &40.21 &30.38    & 45.32 &44.75 &45.41 &44.94     \\
    Gemma-2          & 33.53 &14.84 &34.23 &19.44                      &46.53 &61.53 &46.78 &45.09    & 40.48 &49.79 &40.95 &35.34     \\
    Yi               & 34.44 &35.69 &34.33 &33.73                &42.59 &53.13 &42.95 &39.16    & 45.32 &47.89 &45.23 &45.45     \\
    Mixtral          & 39.27 &19.22 &29.07 &21.33                &55.89 &57.35 &55.64 &54.98    & 48.94 &52.49 &48.69 &47.52     \\
    LLaMA-3.3        & 35.64 &39.65 &36.30 &23.82                      &58.61 &60.24 &58.50 &58.93    & 46.83 &53.03 &47.13 &45.67     \\
    LLaMA-4          & 41.38 &54.51 &41.22 &36.02                &59.00 &61.83 &57.89 &58.72      &46.22  &53.07 &46.22 &46.30\\
    Qwen-3           & 39.27 &39.29 &29.87 &24.11                &48.64 &65.71 &48.80 &45.53      &48.94  &60.55 &49.28 &47.50\\
    DeepSeek-R1      & {\bf 45.61} &53.94 &{\bf 46.06} &{\bf 42.03}                      &50.15 &55.37 &50.46 &49.09    & 50.75 &51.85 &50.99 &49.33     \\    
    Qwen-2.5         & 39.27 &{\bf 56.11} &39.86 &31.27    &{\bf 61.33} &{\bf 67.84} &{\bf 61.22} &{\bf 61.91}    & {\bf 54.38} &60.09 &{\bf 54.47} &{\bf 55.36 }    \\
     \noalign{\hrule height 1.5pt}
    \rowcolor{gray!20}\multicolumn{13}{c}{\it{\textbf{Closed Source General Models}}} \\
    \hline
    GPT-3.5          & 33.23 & 16.18 & 33.92 & 19.17                      & 36.56 & 21.75 & 37.15 & 26.79    & 34.14 & 34.58 & 34.42 & 31.95     \\
    GPT-4            & 33.23 & 14.18 & 33.93 & 18.55                      & 51.36 & 57.39 & 51.68 & 50.18    & 44.11 & 51.16 & 44.51 & 41.51    \\
    GPT-4o           & 34.44 & 17.35 & 35.12 & 22.17                      & 54.38 & 57.88 & 54.59 & 53.97    & 48.94 & 55.34 & 49.26 & 47.87   \\
    Claude-3.5       & 35.95 & 50.92 & 36.59 & 25.37                      & 50.45 & 56.76 & 50.74 & 49.75    & 51.66 & {\bf 56.80} & 51.99 & 50.31     \\
    Doubao-pro       & 41.69 & {\bf 55.81} & 42.17 & 36.35                & 40.79 & 58.51 & 41.24 & 35.75    & {\bf 51.96} & 56.29 & 52.09 & {\bf 52.53}     \\
    ERNIE-3.5        & {\bf 44.11} & 52.55 & {\bf 44.62} & {\bf 39.23}    & 51.66 & 55.47 & 51.89 & 51.32    & 44.79 & 52.81 & 45.54 & 45.25     \\
    Baichuan-4        & 38.97 & 51.85 & 39.57 & 29.91                      & {\bf 56.79} & {\bf 62.39} & {\bf 56.81} & {\bf 57.67}    & {\bf 51.96} & 54.50 & {\bf 52.14} & 51.92     \\
     \noalign{\hrule height 1.5pt}
    \rowcolor{gray!20}\multicolumn{13}{c}{\it{\textbf{Medical Models}}} \\
    \hline
    Medllama          & 31.72 & 10.67 & 32.41 & 16.06                      & 38.07 & 37.84 & 38.16 & 33.74    & 41.39 & 31.32 & 31.13 & 30.73     \\
    HuatuoGPT-o1      & 38.97& 52.44 & 39.48 & 31.85                      & 46.82 & 54.49 & 47.05 & {\bf 46.99}    & {\bf 47.12} & {\bf 58.91} &  {\bf 47.27} & {\bf 45.68}     \\
    Doctorai          & 34.14 & 54.21 & 33.94 & 18.63                      & 42.90 & {\bf 56.33} & {\bf 43.26} & 39.73    & 42.59 & 46.24 &42.72 & 42.68    \\
    Openbiollm        & {\bf 47.73} & {\bf 67.09} & {\bf 48.04} & {\bf 38.85}                      & {\bf 47.13} & 35.99 & 35.56 & 33.08    & 41.39 & 35.76 & 31.38 & 26.89   \\
    \noalign{\hrule height 1.5pt}
\end{tabular}
}
}
}
\label{tab:main_results}
\end{table*}
\begin{table*}[ht]
\centering
\setlength{\tabcolsep}{2mm}        
\caption{The Performance (\%) of the impact of different risk factor compositions in CoT on ROP prediction.}
\vspace{-0.5\intextsep}
{
{
\resizebox{\linewidth}{!}{
\begin{tabular}{l|cccc|cccc|cccc}
    \toprule[1.5pt]
    \multirow{2}{*}{\textbf{Factor}} & 
    \multicolumn{4}{c|}{\textbf{LLaMA-3.3}} & 
    \multicolumn{4}{c|}{\textbf{Qwen-2.5}} & 
    \multicolumn{4}{c}{\textbf{GPT-4o}} \\
    
    \cmidrule(rl){2-5}\cmidrule(rl){6-9} \cmidrule(rl){10-13} 
    & {Acc.} & {Prec.} & {Rec.} & {F1} & {Acc.} & {Prec.} & {Rec.} & {F1} & {Acc.} & {Prec.} & {Rec.} & {F1}\\

    \noalign{\hrule height 1.5pt}
    \rowcolor{gray!20}\multicolumn{13}{c}{\it{\textbf{Risk Factors for ROP}}} \\
    \hline
    add \hyperref[abbreviated]{GA\&W}        & {\bf62.54} & {\bf64.39} & {\bf62.45} & {\bf62.94}                      & 58.91 & 65.27 & 58.75 & 59.02    & {\bf62.54} & {\bf64.03} & {\bf62.52} & {\bf63.03}     \\
    add \hyperref[abbreviated]{MP}       & 58.61 & 58.13 & 58.59 & 58.33                      & 57.40 & 65.16 & 57.19 & 57.13    & 59.82 & 62.01 & 59.84 & 60.45     \\
    add \hyperref[abbreviated]{OE}       & 57.09 & 58.42 & 57.10 & 57.58                      & 58.01 & 65.52 & 57.90 & 58.60    & 55.89 & 58.94 & 56.06 & 55.65     \\
    add \hyperref[abbreviated]{PC}       & 56.49 & 58.39 & 56.56 & 56.97                      & 57.09 & 66.62 & 57.01 & 57.39    & 56.19 & 59.74 & 56.41 & 55.77     \\
    All \hyperref[abbreviated]{Factors}  & 58.61 & 60.24 & 58.50 & 58.93                      & {\bf61.33} & {\bf67.84} & {\bf61.22} & {\bf61.91}    & 54.38 & 57.87 & 54.59 & 53.97     \\
    \hline
\end{tabular}
}
}
}
\label{tab:factor}
\end{table*}

\subsection{\!ROP Risk Prediction Capabilities of LLMs}
\noindent{\bf Performance of Instructional Scheme:}
First, we verified whether leveraging only the Instruction scheme enables LLMs to predict ROP risk? 
The experimental results, as shown in Table \ref{tab:main_results}, indicate that the majority of LLMs struggle to predict ROP risk, resulting in lower accuracy.
For instance, the accuracy of the LLaMA-3.3 model is 35.64\%, which suggests that relying solely on the knowledge of LLMs is ineffective for predicting ROP risk and providing effective diagnostic assistance.
Furthermore, as the number of model parameters increases, the DeepSeek-R1 model achieves the best performance in the open-source general models setting. Additionally, compared to the LLaMA-3.3 model, the distilled DeepSeek-R1 model shows a significant improvement, with an increase in accuracy of 9.97\%.

\noindent{\bf Performance of CoT Scheme:}
We verify the performance of the CoT scheme on LLMs.
It is evident that, upon introducing the CoT, the prediction accuracy of the LLMs significantly improved compared to the Instructional scheme.
For example, the LLaMA-3.3 shows an average improvement of 25.21\%.
Simultaneously, the Qwen-2.5 and Baichuan-4 models achieve the best performance in the open source and closed source model settings, respectively.
The results above indicate that the CoT scheme helps to enhance model predictive performance and promotes the alignment of reasoning capabilities with known risk factors in ROP.

\begin{figure*}[t] 
  \centering
\includegraphics[width=1.02 \textwidth]{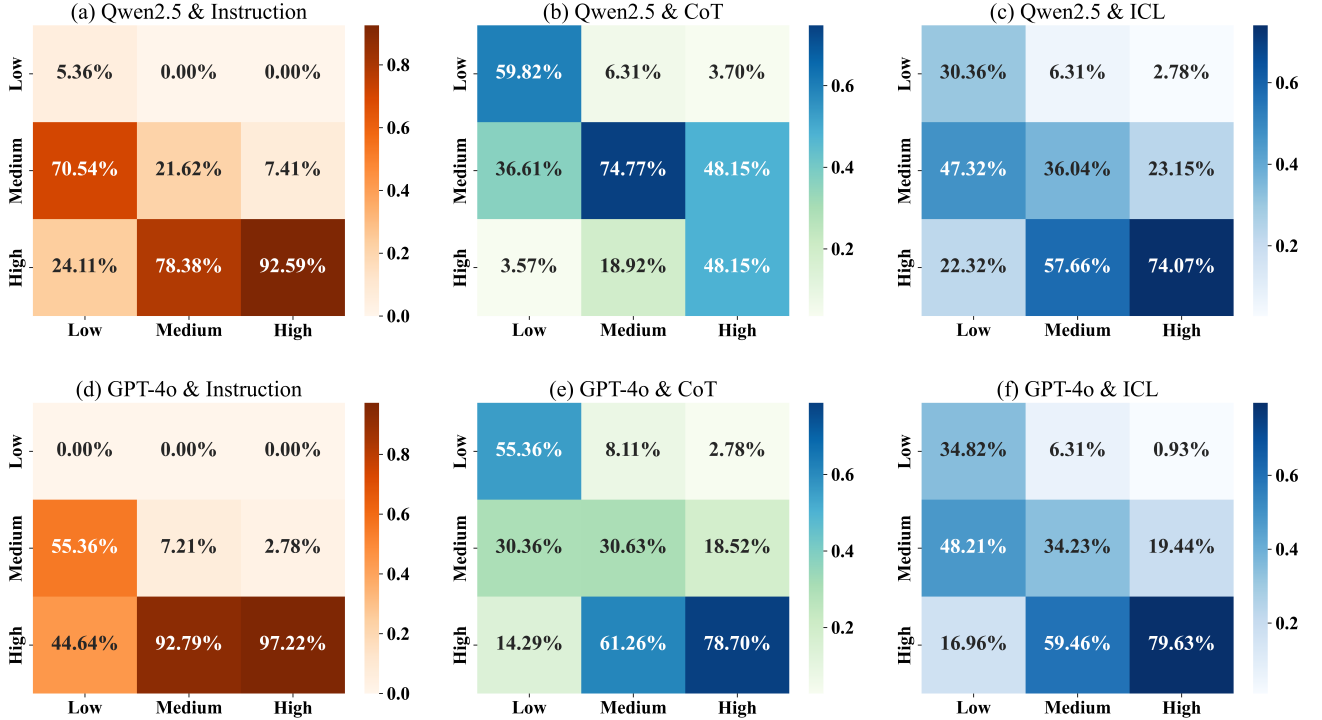}
\caption{The confusion matrix of the model's predicted risk compared with the true label. The $x$-axis represents the model predictions, while the $y$-axis represents the true labels.}%
\label{fig2}
\end{figure*}

\noindent{\bf Performance of ICL Scheme:}
For the ICL scheme, the prediction accuracy of the LLMs for ROP risk also improves compared to the Instructional scheme.
For instance, the average performance improves by 15.48\% in the Claude-3.5.
However, although the ICL scheme improves model performance, it is not as effective as the CoT scheme. For example, in the Qwen-2.5, the average performance decreases by 7\%.
The reason for this is that, compared to the CoT scheme composed of known risk factors, the ICL scheme, made up of demonstration examples, more rigorously challenges the model's reasoning abilities and its intrinsic understanding of ROP disease.

\noindent{\bf Performance Comparison for Different Source Models:}
From another perspective, we evaluate the performance of different source models. 
In the Instructional scheme, the medical model Openbiollm achieves the best performance, with an ROP risk prediction accuracy of 47.73\%, which indicates that models fine-tuned with more specialized medical samples can achieve more precise ROP diagnostic capabilities.
For instance, the accuracy improves by 12.09\% compared to the LLaMA-3.3.
However, in the CoT and ICL schemes, the open-source Qwen-2.5 outperforms other LLMs.
The potential reason may be that Qwen-2.5 consistently outperforms other models in Chinese tasks, which aligns with the description by Qwen-2.5~\cite{qwen2.5}.


Although the CoT and ICL schemes help to improve the model's accuracy in predicting ROP risk, the best-performing Qwen-2.5 model only achieves an accuracy of 61.33\%, which is not considered ideal performance.
Therefore, through the validation of Affective-ROPTester's three schemes, our conclusion for Question \hyperref[ques:effectiveness]{1} is that the current LLMs are not sufficiently effective at predicting ROP risk.

\begin{table*}[t]
\centering
\setlength{\tabcolsep}{2mm}        
\caption{The performance (\%) of the impact of \textbf{different affective framings} on ROP prediction.}
\vspace{-0.5\intextsep}
{
{
\resizebox{\linewidth}{!}{
\begin{tabular}{c|cccc|cccc|cccc}
    \toprule[1.5pt]
    \textbf{Sample} & 
    \multicolumn{4}{c|}{\textbf{Negative}} & 
    \multicolumn{4}{c|}{\textbf{Neutral}} & 
    \multicolumn{4}{c}{\textbf{Positive}} \\
    
    \cmidrule(rl){2-5}\cmidrule(rl){6-9} \cmidrule(rl){10-13} 
\textbf{Number} & {Acc.} & {Prec.} & {Rec.} & {F1} & {Acc.} & {Prec.} & {Rec.} & {F1} & {Acc.} & {Prec.} & {Rec.} & {F1}\\

    \noalign{\hrule height 1.5pt}
    \rowcolor{gray!20}\multicolumn{13}{c}{\it{\textbf{Different Affective Framings}}} \\
    \hline
    LLaMA-3.3        & 33.53 & 16.22 & 34.22 & 19.40        & 37.76 & 46.13 & 38.40 & 27.76    & 50.45 & 47.75 & 50.73 & 46.25     \\
    Qwen-2.5         & 37.76 & 53.94 & 38.39 & 28.31        & 40.78 & 57.62 & 41.30 & 34.36    & 42.29 & 54.86 & 42.72 & 38.70     \\
    DeepSeek-R1     & 40.18 & 46.34 & 40.77 & 32.16        & 43.50 & 51.57 & 43.98 & 39.11    & 49.24 & 54.18 & 49.58 & 47.57     \\
    o4-mini         & 49.54 & 40.79 & 37.44 & 35.03        & {\bf54.98} & {\bf60.64} & {\bf55.30} & 53.28    & 53.47 & 57.18 & 53.69 & 52.89     \\
    o3-mini         & {\bf50.75} & {\bf57.50} & {\bf51.09} & {\bf49.35}        & 54.07 & 59.23 & 54.30 & {\bf53.99}    & {\bf57.09} & {\bf58.69} & {\bf57.18} & {\bf57.42}     \\
    \hline
\end{tabular}
}
}
}
\label{tab:affective}
\end{table*}

\begin{figure*}[ht] 
  \centering
\includegraphics[width=1.0 \textwidth]{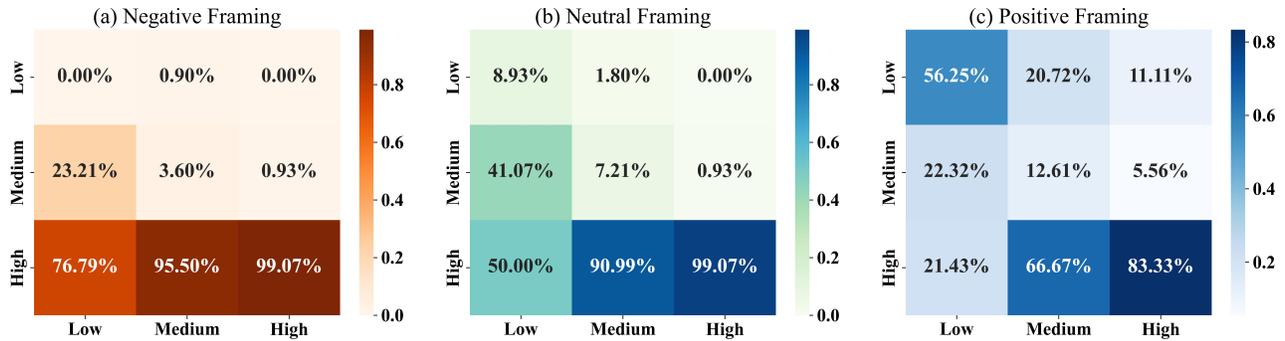}
\caption{The confusion matrix of the predicted risk under \textbf{different affective framings} compared with the true labels.}%
\label{fig_affective}
\end{figure*}

\begin{table*}[ht]
\centering
\setlength{\tabcolsep}{2mm}        
\caption{The performance (\%) of the impact of different demonstration examples number on ROP prediction.}
\vspace{-0.5\intextsep}
{
{
\resizebox{\linewidth}{!}{
\begin{tabular}{c|cccc|cccc|cccc}
    \toprule[1.5pt]
    \textbf{Sample} & 
    \multicolumn{4}{c|}{\textbf{LLaMA-3.3}} & 
    \multicolumn{4}{c|}{\textbf{Qwen-2.5}} & 
    \multicolumn{4}{c}{\textbf{GPT-4o}} \\
    
    \cmidrule(rl){2-5}\cmidrule(rl){6-9} \cmidrule(rl){10-13} 
\textbf{Number} & {Acc.} & {Prec.} & {Rec.} & {F1} & {Acc.} & {Prec.} & {Rec.} & {F1} & {Acc.} & {Prec.} & {Rec.} & {F1}\\

    \noalign{\hrule height 1.5pt}
    \rowcolor{gray!20}\multicolumn{13}{c}{\it{\textbf{Number of the Demonstration Examples}}} \\
    \hline
    One       & 42.59 & 50.47 & 43.14 & 36.21                      & 50.15 & 58.47 & 50.42 & 49.24    & 43.50 & 57.39 & 43.94 & 39.52     \\
    Two       & 41.69 & 51.63 & 42.13 & 38.11                      & 51.06 & 56.99 & 51.29 & 50.96    & 46.53 & 59.33 & 46.91 & 44.15     \\
    Three     & 45.02 & 51.17 & 45.40 & 42.55                      & 52.27 & 58.89 & 52.39 & 53.09    & 45.92 & 57.21 & 46.26 & 44.70     \\
    Four      & 45.92 & 52.89 & 46.27 & 44.27                      & 54.08 & 59.65 & 54.17 & 55.16    & {\bf 50.15} & {\bf 60.29} & {\bf 50.48} & {\bf 49.01}     \\
    Five      & {\bf 46.83} & {\bf 53.03} & {\bf 47.13} & {\bf 45.67}                      & {\bf 54.38} & {\bf 60.09} & {\bf 54.47} & {\bf 55.36}    & 48.94 & 55.34 & 49.26 & 47.87     \\
    \hline
\end{tabular}
}
}
}
\label{tab:number}
\end{table*}

\subsection{Bias in ROP Risk Prediction of LLMs}\label{bias_llm}
To verify whether LLMs have an intrinsic bias in predicting ROP risk, we analyze the confusion matrix of the risk outputs from the Qwen-2.5 and GPT-4 models, as shown in Figure \ref{fig2}. 
Specifically, in the Instructional scheme (first column), we observe that the model tends to output high or medium risk. For example, in the Qwen-2.5 model, the prediction accuracy for low-risk samples is only 5.36\%, while the accuracy for high-risk samples is 92.59\%.
In the case of medium-risk samples, the model also tends to predict high-risk and does not include any low-risk predictions.

Compared to the Qwen-2.5 model, the GPT-4o model shows a more pronounced tendency to predict high risk. For example, in low-risk samples, the accuracy of correct predictions is 0\%, and the predictions for high risk make up 44.64\% of the total samples, which is significantly higher than the 24.11\% in the Qwen-2.5 model.
This situation also occurs in medium-risk samples, where the GPT-4o model predicts medium risk as high risk at a rate of 92.79\%.
The analysis above indicates that naive LLMs tend to predict ROP risk as medium or higher, which answers Question \hyperref[ques:bias]{2}, confirming that LLMs exhibit bias in predicting ROP risk.
Furthermore, the GPT-4o exhibits a more pronounced bias compared to the Qwen-2.5, which is the reason why the performance of the GPT-4o model is less robust than that of the Qwen-2.5.

Additionally, in the CoT and ICL schemes, the bias in the model is significantly mitigated. 
The accuracy of predicting low-risk samples has greatly increased, and the proportion predicted as high-risk has decreased.
For example, in low-risk samples, the Qwen-2.5 model outputs low-risk predictions at a rate of 59.82\%, and the proportion of high-risk predictions has decreased to 3.57\% under the CoT scheme.
This indirectly reconfirms that relying solely on the model's internal knowledge to predict ROP risk is biased.

\begin{figure*}[ht] 
  \centering
\includegraphics[width=1.0 \textwidth]{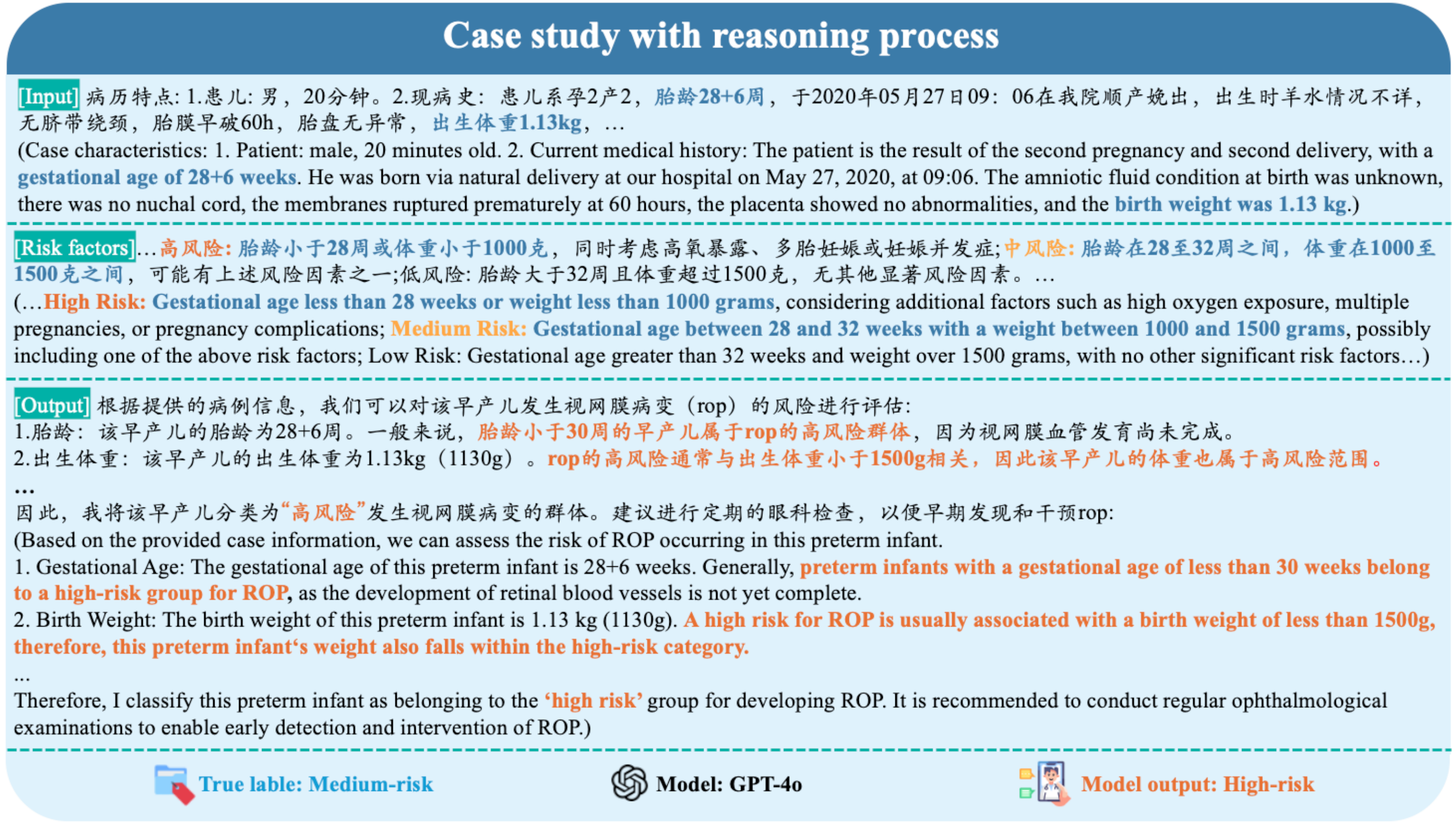}
\caption{A case study with ROP risk prediction reasoning process based on Affective-ROPTester's Instructional scheme, the model used is GPT-4o.}%
\label{fig3}
\end{figure*}

Finally, we present a case study illustrating the diagnostic process in GPT-4o.
As shown in Figure \ref{fig3}, the preterm infant with a gestational age of nearly 29 weeks and a birth weight of 1.13 kg meets criteria for medium-risk according to known risk factors. However, the GPT-4o incorrectly classifies infants with gestational ages under 30 weeks and weights below 1.5 kg as high-risk, which is a clear error. 
Therefore, the reason for the bias in the model may be its excessive rigor, leading it to favor high-risk predictions.

\subsection{ROP Risk Prediction with Different Affective Framings}
\noindent{\bf Performance Comparison of Different Affective Framings}:
As shown in Table \ref{tab:affective}, different affective framings have a significant impact on the model's ability to predict ROP risk. We observe that, compared to neutral and positive affective framings, negative affective framing has a notably adverse effect on prediction accuracy. For example, in the DeepSeek-R1 model, the accuracy under negative affective framing is 40.18\%, while under neutral affective framing it rises to 43.50\%, showing an improvement of 3.32\%. Furthermore, the accuracy under positive affective framing reaches 49.24\%, representing a further increase of 9.06\%. This pattern is also observed across other models. Therefore, we conclude that, compared to negative affective framing, neutral or positive affective framing contributes to improved accuracy in ROP risk prediction, addressing Question \hyperref[ques:affective]{3}.

\noindent{\bf Impact of Different Affective Framings on Bias}:
In Section \ref{bias_llm}, we observed that LLMs tend to predict medium and high risk for ROP, which may be attributed to the model's overly pessimistic assumptions about premature infants. A plausible solution is to incorporate affective prompts to mitigate biased predictions. As shown in Figure \ref{fig_affective}, we present the model's risk prediction tendencies under different affective framings. It is evident that under negative and neutral affective framings, the model tends to predict medium and high risk, which is consistent with the observations in Section \ref{bias_llm}.
In contrast, when exposed to different affective framings, this bias is significantly mitigated. For example, when the input sample corresponds to a low-risk case, the probability of the model predicting high risk is 76.79\% under the negative affective framing, 50\% under the neutral affective framing, and drops to only 21.43\% under the positive affective framing. This indicates that, in the context of ROP risk prediction, prompts infused with positive affective framing help reduce the bias exhibited by LLMs.

\subsection{Ablation Experiments of Affective-ROPTester}
\noindent{\bf Different Risk Factors for ROP: }To analyze the impact of known risk factors on ROP prediction in the CoT scheme, we conduct ablation experiments.
From Table \ref{tab:factor}, we find that when using only gestational age and birth weight for the CoT, the optimal ROP risk prediction occurs in the LLaMA-3.3 and GPT-4o models.
For example, the average metric is 63.03\% in the GPT-4o model, which is higher than the results of other risk factors.
The reason for this phenomenon lies in the fact that, compared to other risk factors, gestational age and birth weight possess well-defined numerical foundations, facilitating more effective assistance in model predictions.

\begin{table*}[t]
\centering
\setlength{\tabcolsep}{2mm}        
\caption{The performance (\%) of the impact of different large language model size on ROP risk prediction.}
{
{
\resizebox{\linewidth}{!}{
\begin{tabular}{l|cccc|cccc|cccc}
    \toprule[1.5pt]
    \multirow{2}{*}{\textbf{Model}} & 
    \multicolumn{4}{c|}{\textbf{Instruction}} & 
    \multicolumn{4}{c|}{\textbf{Chain of Thought}} & 
    \multicolumn{4}{c}{\textbf{ICL}} \\
    
    \cmidrule(rl){2-5}\cmidrule(rl){6-9} \cmidrule(rl){10-13} 
    & {Acc.} & {Prec.} & {Rec.} & {F1} & {Acc.} & {Prec.} & {Rec.} & {F1} & {Acc.} & {Prec.} & {Rec.} & {F1}\\

    \noalign{\hrule height 1.5pt}
    \rowcolor{gray!20}\multicolumn{13}{c}{\it{\textbf{Performance of LLMs with Different Sizes}}} \\
    \hline
    Qwen2.5:0.5B         & 36.25 & 25.91 & 35.75 & 23.46                      & 36.86 & 27.52 & 36.35 & 24.05    & 38.97 & 51.69 & 38.49 & 29.31     \\
    Qwen2.5:1.5B         & 34.44 & 34.12 & 34.63 & 33.08                      & 32.93 & 22.12 & 33.62 & 18.14    & 37.46 & 39.59 & 37.29 & 36.23     \\
    Qwen2.5:3B           & 35.65 & 38.35 & 35.22 & 23.48                      & 45.32 & 50.11 & 44.95 & 40.34    & 43.50 & 45.43 & 43.39 & 43.25     \\
    Qwen2.5:7B           & 34.74 & 48.04 & 35.41 & 21.78                      & 57.40 & 58.57 & 57.63 & 55.34    & 42.29 & 53.63 & 42.36 & 41.47     \\
    Qwen2.5:14B          & {\bf 43.50} & 55.19 & {\bf 43.77} & {\bf 39.09}                      & 53.47 & 65.48 & 53.47 & 52.91    & 47.73 & 57.33 & 47.97 & 47.53     \\
    Qwen2.5:32B          & 38.97 & 55.20 & 39.54 & 31.72                      & {\bf 61.63} & 64.34 & {\bf 61.63} & {\bf 62.36}    & 52.26 & 58.35 & 52.42 & 52.87     \\
    Qwen2.5:72B          & 39.27 & {\bf 56.11} & 39.86 & 31.27                      & 61.33 & {\bf 67.84} & 61.22 & 61.91    & {\bf 54.38} & {\bf 60.09} & {\bf 54.47} & {\bf 55.36}     \\    
     \hline
\end{tabular}
}
}
}
\label{tab:size}
\end{table*}

\begin{figure*}[!t]
  \centering
  \captionsetup[subfloat]{font=scriptsize}
  \subfloat[Instruction scheme]{\includegraphics[width=2.2in]{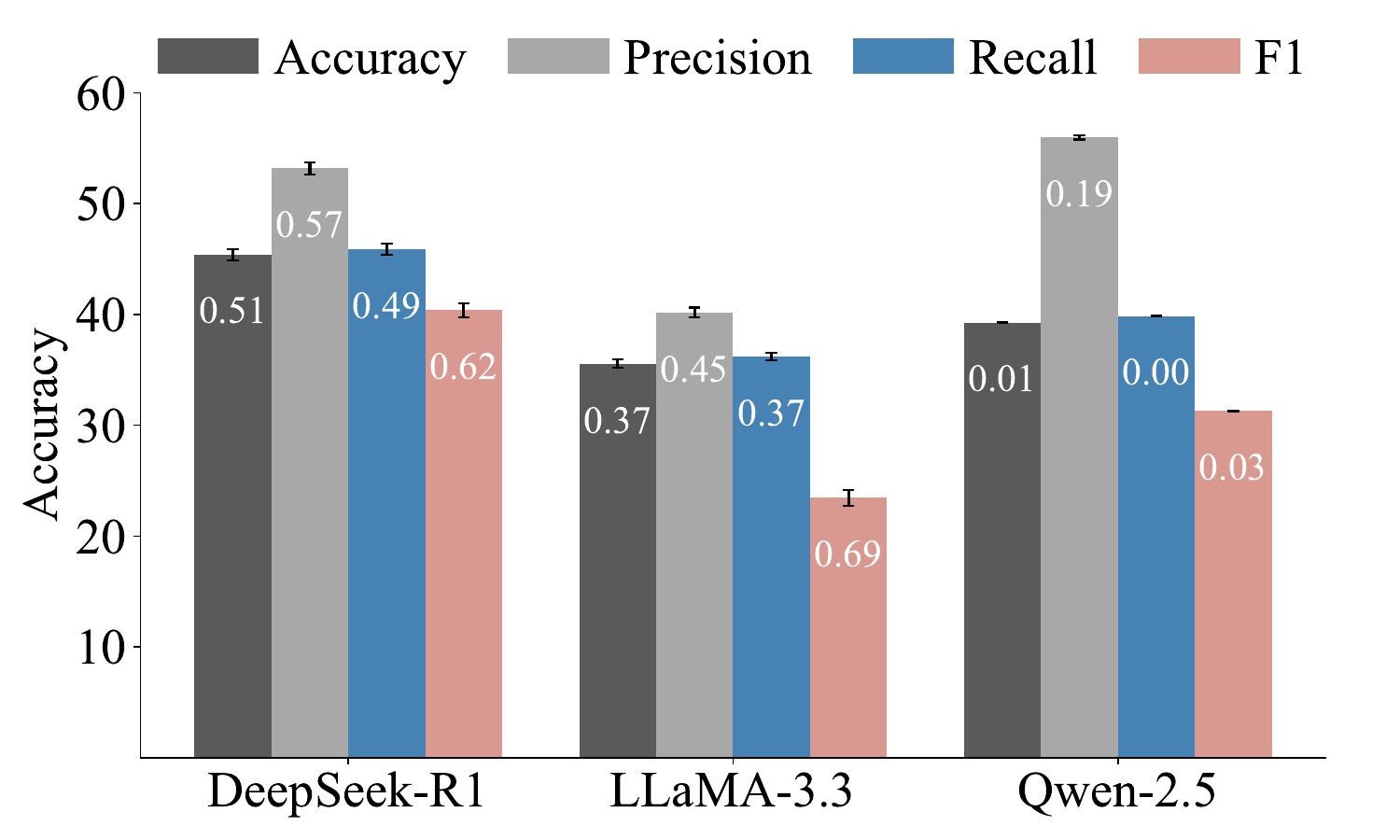}\label{fig:6.1}}
  \subfloat[CoT scheme]{\includegraphics[width=2.2in]{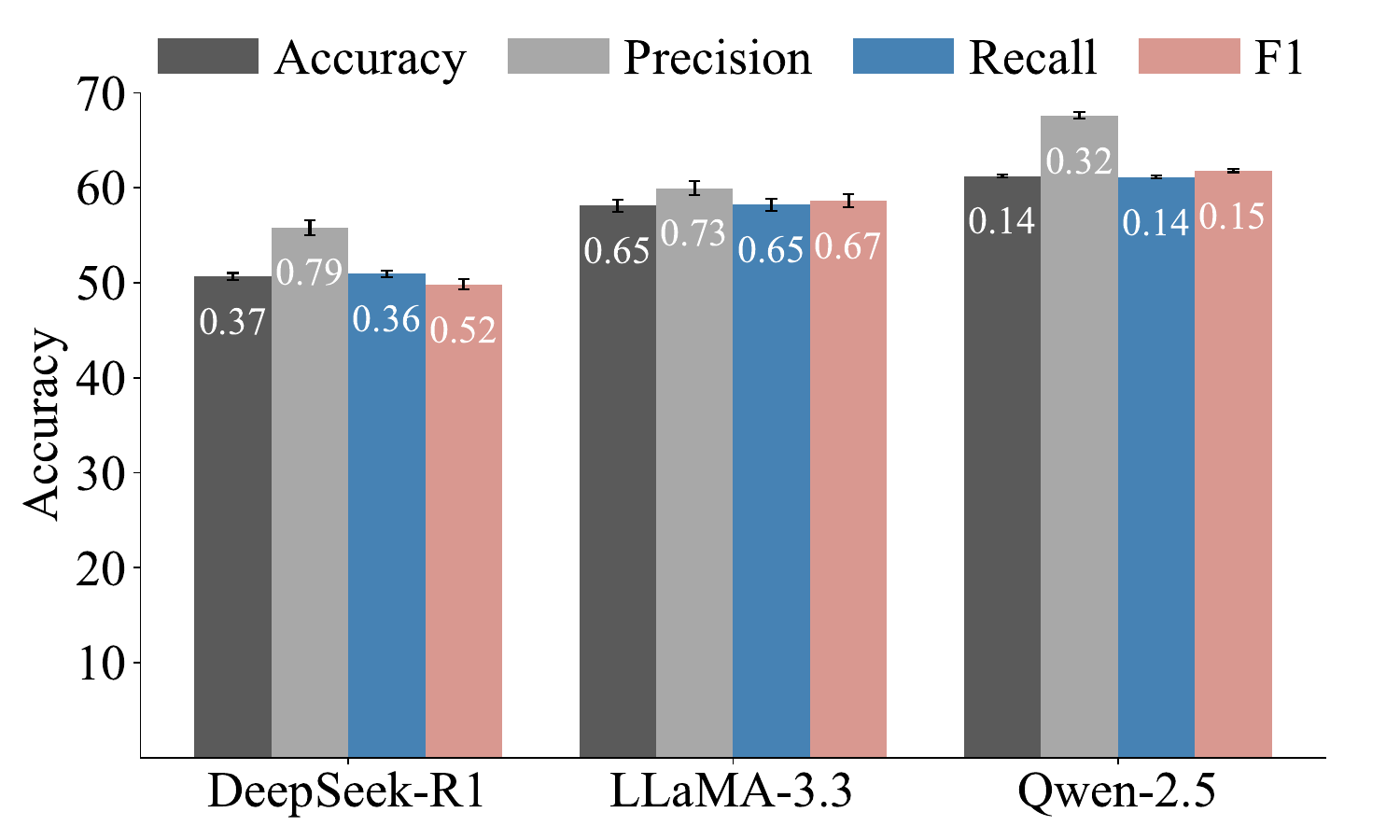}\label{fig:6.2}}
  \subfloat[ICL scheme]{\includegraphics[width=2.2in]{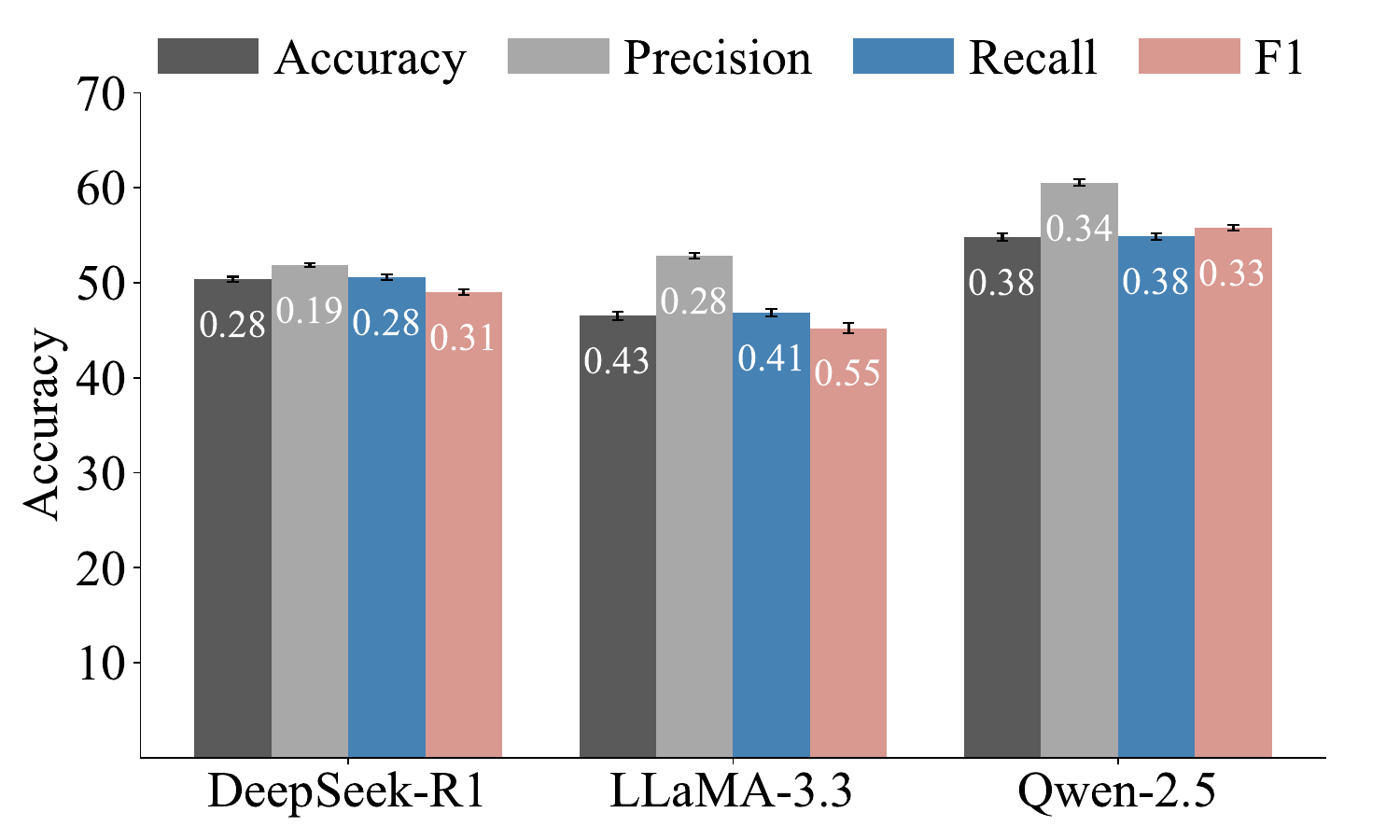}\label{fig:6.3}}
\caption{The predicted results with {\bf standard deviation} under the Instruction, CoT, and ICL schemes.}
\label{fig:prompt} 
\end{figure*}

\noindent{\bf Different Numbers of Demonstration Examples for ROP Risk:}
We analyze the impact of varying numbers of demonstration examples on the ICL scheme. \!The results are shown in Table \ref{tab:number}. 
As the number of demonstration examples increases, the model's accuracy in predicting ROP risk gradually improves.
For example, the average metric increased from an initial 43.1\% to 48.16\% in the LLaMA-3.3, which indicates that demonstration examples can assist the model in predicting ROP risk.
Additionally, the model performance is optimal when the number of demonstration examples is four in the GPT-4o model, indicating that an excessive number of demonstration examples may also affect prediction accuracy.

\begin{table*}[ht]
\centering
\setlength{\tabcolsep}{2mm}        
\caption{Performance analysis (\%) when the dataset categories are unbalanced.}
{
{
\resizebox{\linewidth}{!}{
\begin{tabular}{c|cccc|cccc|cccc}
    \toprule[1.5pt]
    \multirow{2}{*}{\textbf{Model}} & 
    \multicolumn{4}{c|}{\textbf{Instruction}} & 
    \multicolumn{4}{c|}{\textbf{Chain of Thought}} & 
    \multicolumn{4}{c}{\textbf{ICL}} \\
    
    \cmidrule(rl){2-5}\cmidrule(rl){6-9} \cmidrule(rl){10-13} 
    & {Acc.} & {Prec.} & {Rec.} & {F1} & {Acc.} & {Prec.} & {Rec.} & {F1} & {Acc.} & {Prec.} & {Rec.} & {F1}\\

    \noalign{\hrule height 1.5pt}
    \rowcolor{gray!20}\multicolumn{13}{c}{\it{\textbf{Performance of LLMs with an Unbalanced Dataset.}}} \\
    \hline
    Qwen-2.5     & 19.01 & 42.86 & 42.41 & 20.11   & 53.90 & {\bf54.92} & {\bf55.06} & 49.25    & 45.01 & 48.62 & 55.32 & 42.59     \\
    Qwen-3       & 17.70 & 30.43 & 29.35 & 14.09   & 29.94 & 40.03 & 36.82 & 26.19    & 32.32 & 34.04 & 37.73 & 24.41     \\
    LLaMA-3.3    & 20.31 & 39.99 & 39.30 & 17.88   & 56.25 & 49.91 & 51.28 & 47.16    & 51.35 & 46.70 & 57.81 & 42.68     \\
    LLaMA-4      & 30.98 & {\bf53.71} & 42.30 & 29.66   & {\bf57.81} & 53.85 & 55.04 & {\bf50.77}    & 47.12 & 47.45 & 54.41 & 42.34     \\
    o3-mini      & 44.53 & 46.69 & {\bf54.74} & {\bf40.18}   & 44.27 & 53.55 & 52.62 & 44.68    & 53.17 & 46.72 & {\bf58.64} & 44.42    \\
    o4-mini      & {\bf46.35} & 44.46 & 52.65 & 38.24   & 42.44 & 54.33 & 53.71 & 44.08    & {\bf56.19} & {\bf47.60} & 57.86  & {\bf45.78}     \\
     \hline
\end{tabular}
}
}
}
\label{tab:unblanced}
\end{table*}

\noindent{\bf Reasoning and Fine-tuning:}\label{Rea} 
Table \ref{tab:thinking} demonstrates whether the model with reasoning can optimize the performance of ROP diagnosis.
Firstly, we found that compared to the DeepSeek-R1 reasoning model, the performance of the o3-mini significantly improved, with accuracy increasing by 6.64\%.
Additionally, compared to the non-reasoning model GPT-4o, accuracy increased by 2.41\%, which indicates that reasoning contributes to enhancing the model's performance in predicting ROP.
However, compared to the Qwen-2.5 model, the performance of the reasoning models is still lacking.
Finally, we present a case study of the process for predicting ROP in Figure \ref{fig3}.
It is not difficult to observe that the model classifies preterm infants with a gestational age greater than 28 weeks as high-risk, which does not match the known risk factors.
This prediction process once again demonstrates the model's inadequacy in predicting ROP.
In addition, we fine-tuned the Qwen2.5:7B model leveraging LoRA~\cite{hulora}. It is not difficult to see that the prediction accuracy of the Qwen2.5:7B model is only 44.11\%.
\begin{table}[H]
\centering
\setlength{\tabcolsep}{1.0mm}   
\caption{Experimental results (\%) of the model with reasoning and fine-tuning.}
{
{
\renewcommand{\arraystretch}{1.0}\resizebox{0.385 \textwidth}{!}{\begin{tabular}{c|c|cccc}
\toprule[1.5pt]
    Method & Model & {Acc.} & {Prec.} & {Rec.} & {F1} \\
\hline
   Reasoning     &o3-mini   & 56.79 & 65.32 & 56.78 & 56.88  \\
   Reasoning     &o4-mini   & 54.07 & 63.60 & 54.12 & 54.11  \\
   Reasoning     &o1        & 55.89 & 64.38 & 55.86 & 56.07  \\
\hline
   Fine-tuning       & Qwen2.5:7B      & 44.11 & 45.30 & 43.89 & 43.13  \\
\hline
\end{tabular}
}}
}
\label{tab:thinking}
\end{table}

\noindent{\bf Models of Different Sizes for ROP:}
To further verify the ability of LLMs with different parameter sizes to predict ROP risk, we conduct experiments using the Qwen-2.5 series of models, which range from 0.5B to 72B parameters.
As shown in Table \ref{tab:size}, we observe that in the Instructional scheme, the accuracy of model predictions for ROP risk is not positively correlated with the size of the model parameters.
This is because relying solely on the intrinsic knowledge of the model is insufficient for an accurate diagnosis of ROP risk.
In the CoT and ICL schemes, the incorporation of external knowledge significantly enhances the accuracy of ROP risk predictions. 
Furthermore, with an increase in the model's parameter size, there is a corresponding enhancement in its reasoning capabilities, which in turn progressively elevates the accuracy of ROP risk predictions. 

\noindent{\bf Results of Different Structures:} \label{c.1}
By comparing the performance of models with different network structures, we found that models with a dense structure perform better than those using a mixture of experts (MoE) model.
As shown in Table \ref{tab:aoe}, the Qwen2.5:7B model exhibits a 1.51\% improvement in prediction accuracy compared to the Mixtral model.
In addition, we also explore the model's performance for ROP risk predictions within the setting that combines the CoT and ICL schemes. Experimental results show that this did not yield improvements; instead, the accuracy decreased by 4.84\% in the Qwen2.5:72B model.

\begin{table}[h]
\centering
\setlength{\tabcolsep}{1.0mm} 
\caption{The comparison of performance (\%) under different structures and strategies.}
{{\renewcommand{\arraystretch}{1.0}\resizebox{0.43 \textwidth}{!}{
\begin{tabular}{c|c|cccc}
\toprule[1.5pt]
    Method & Model & {Acc.} & {Prec.} & {Rec.} & {F1} \\
\hline
   MoE           & Mixtral    & 55.89 & 57.35 & 55.64 & 54.98  \\
   MoE           & GPT-4o      & 54.38 & 57.88 & 54.59 & 53.97  \\
\hline
 Dense           & LLaMA3.3:70B    & 58.61 & 60.24 & 58.50 & 58.93  \\
 Dense           & Qwen2.5:7B      &  57.40 & 58.57 & 57.63 & 55.34 \\
 Dense           & Qwen2.5:72B      & 61.33 & 67.84 & 61.22 & 61.91 \\
\hline
 CoT\&ICL         & Qwen2.5:72B      & 56.49 & 60.41 & 56.48 & 57.52  \\
\toprule[1.0pt]
\end{tabular}}}}
\label{tab:aoe}
\end{table}

\noindent{\bf Results with Standard Deviation:} \label{c.2}
Furthermore, we present the predicted results with standard deviation in the DeepSeek-R1, LLaMA-3.3, and Qwen-2.5 models. As shown in Figure \ref{fig:prompt}, we observe that the standard deviations for all metrics are mostly less than 0.5. For example, in the ICL scheme setting, the standard deviation for the DeepSeek-R1 model is 0.28, which indicates that our experimental results have feasible stability.

\noindent{\bf Results of the Unbalanced Dataset:}\label{c.7}
To further demonstrate the reliability of our findings, such as the potential bias in the model's prediction of ROP, we conduct experiments with all the samples (totaling 3,922) without considering the balance of sample categories.
The unbalanced dataset includes 2,813 low-risk samples, 848 medium-risk samples, and 331 high-risk samples.
The experimental results are shown in Table \ref{tab:unblanced}. We found that the accuracy of the model's predictions is lower compared to the results in Table \ref{tab:main_results}. For example, the accuracy of the Qwen-2.5 model is only 19.01\% in the Instruction scheme.
The reason is that the model has a bias towards predicting ROP as high-risk or medium-risk, while the unbalanced dataset contains a larger number of low-risk samples, which leads to a higher proportion of incorrect predictions.
The results mentioned above further validate our findings that the model has a bias toward predicting ROP.

\noindent{\bf Disease Prediction vs. Disease Diagnosis:} \label{c.6}
In the domain of traditional smart healthcare, LLMs may be employed to diagnose whether patients have diseases based on medical records and imaging examinations. This approach is utilized to investigate the current health status of patients.
For example, in the case of ROP, LLMs are capable of diagnosing whether premature infants have developed retinal pathology~\cite{sankari2023automated}.
However, our research differs from traditional ROP diagnosis. Instead, it focuses on predicting the risk of ROP in premature infants, based solely on admission notes.
Our motivation lies in predicting the risk of ROP in premature infants in advance, under the premise of an unclear pathogenesis of ROP, with the goal of early detection and treatment.

\begin{table*}[htb]
\centering
\caption{The output of the GPT-4 model presents an introduction to knowledge about ROP.}
\vspace{-1.0\intextsep}
\begin{tcolorbox}[width=0.95\linewidth, boxsep=0pt, left=2pt, right=2pt, top=1pt, bottom=0pt, colback=white, after=\vspace{-3pt}]
\setlength{\baselineskip}{10pt}
{\bf \small Introduce Retinopathy of Prematurity (ROP):} \strut 

\small {\bf GPT-4 model:} 

Retinopathy of Prematurity (ROP) is a potentially blinding eye disorder primarily affecting premature infants. This condition is associated with abnormal development of the retinal blood vessels in the eyes of these infants. The smaller and more premature the infant, the higher the risk and potential severity of ROP.

{\bf Understanding ROP:}

Etiology and Risk Factors:

ROP typically occurs in both eyes and is most common in infants born before 31 weeks of gestation or those weighing less than 1250 grams (about 2.75 pounds) at birth. The development of ROP is influenced by:

\quad $\bullet$ Prematurity: The level of prematurity is directly proportional to the risk of developing ROP.

\quad $\bullet$ Oxygen Exposure: High levels of oxygen supplementation used in neonatal care can also increase the risk of ROP, as oxygen can disrupt the normal development of blood vessels in the retina.

\quad $\bullet$ Other factors include anemia, blood transfusions, and overall health status of the infant.

{\bf Pathophysiology:}

In premature infants, the retinal blood vessels may not have developed fully before birth. Post-birth, these vessels continue to develop, but in some cases, the process becomes abnormal. Initially, vessel growth ceases (phase I) and then proceeds excessively (phase II), leading to the proliferation of abnormal blood vessels. These abnormal vessels are frail and prone to leaking, which can lead to scarring and retinal detachment, the main causes of vision loss in ROP.

{\bf Stages of ROP:}

ROP is classified into five stages, ranging from mild (stage 1) to severe (stage 5):

\quad $\bullet$ Stage 1: Mildly abnormal blood vessel growth.

\quad $\bullet$ Stage 2: Moderately abnormal growth.

\quad $\bullet$ Stage 3: Severe abnormal blood vessel growth. Neovascularization occurs, and there is a risk of the retina starting to detach.

\quad $\bullet$ Stage 4: Partial retinal detachment.

\quad $\bullet$ Stage 5: Total retinal detachment and a high likelihood of blindness.

{\bf Screening and Diagnosis:}

Because the early stages of ROP are typically asymptomatic, regular eye examinations are crucial for premature infants. These exams begin a few weeks after birth and are performed by ophthalmologists using specialized tools to visually inspect the retina.

{\bf Treatment:}

Treatment options depend on the severity of ROP and include:

\quad $\bullet$ Laser Therapy: Used to treat abnormal blood vessels and prevent them from growing further.

\quad $\bullet$ Cryotherapy: A method where extreme cold is used to treat abnormal retinal tissue.

\quad $\bullet$ Anti-VEGF Injections: These are medications that inhibit the growth factors driving the abnormal blood vessel development.

\quad $\bullet$ Surgical Options: In advanced cases, surgeries like scleral buckling or vitrectomy might be necessary to repair retinal detachment.

{\bf Prevention and Management:}

Preventive measures include careful monitoring and adjustment of oxygen levels in premature infants and timely screening for ROP. Early detection and appropriate treatment are crucial in preventing the progression of the disease and preserving vision.

Understanding and managing ROP is a critical aspect of neonatal care and pediatric ophthalmology, emphasizing the need for specialized care and regular monitoring of at-risk infants to prevent serious long-term visual impairment.
\strut
\end{tcolorbox}
\label{tab_introduce_gpt4}
\end{table*}

\noindent{\bf Model's Knowledge about ROP:} \label{c.8}
In Table \ref{tab_introduce_gpt4}, we present the GPT-4 model's introduction to ROP. It is evident that the model possesses critical knowledge of ROP, such as leveraging weight and gestational age as risk factors for predicting ROP.
Although the model demonstrates knowledge of ROP, the experimental results indicate that its ability to predict ROP is insufficient and significantly biased, underscoring the importance of further exploration into ROP prediction algorithms.

\section{Conclusion}
In this paper, we propose the Affective-ROPTester, a comprehensive framework that comprises Instruction, CoT, and ICL schemes, to fully elicit the predictive capabilities of LLMs for ROP risk. We further integrate different affective framings into the prompts to systematically investigate the influence of emotional elements on the models' ability to predict ROP risk. To support these explorations, we construct a new benchmark dataset, named CROP, specifically targeted at ROP risk.
Through extensive experiments with Affective-ROPTester, we demonstrate that LLMs employing the Instruction scheme cannot effectively predict the risks associated with ROP.
Besides, the CoT and ICL schemes can effectively enhance the performances, especially for open-source LLMs.
Furthermore, we observe that LLMs exhibit significant biases in predicting ROP, especially in predicting low-risk samples as medium or high risk, and these biases could be mitigated by introducing external knowledge. 
Finally, incorporating positively framed emotional prompts can enhance the accuracy of ROP risk prediction and help mitigate biases in LLMs.
We plan to build an ROP-specified agent and import a pediatric clinical medicine knowledge database to solve this problem.

\normalem
\bibliography{custom}
\bibliographystyle{IEEEtran}

\vfill

\end{document}